\documentclass{article}
\usepackage{spconf,amsmath,graphicx}
\usepackage{subfigure}
\usepackage{amssymb}
\usepackage{graphicx}
\usepackage{tabularx}
\usepackage{array}

\usepackage[absolute,showboxes]{textpos}
\usepackage{tikz}
\usepackage{lipsum}

\usepackage{ tipa }
\usepackage{ragged2e}
\newcolumntype{P}[1]{>{\RaggedRight\hspace{0pt}}p{#1}}
\newcolumntype{C}[1]{>{\centering}m{#1}}
\usepackage{epstopdf}
\usepackage[headsepline]{scrpage2}
\clearscrheadfoot
\ohead{Initial version of the paper accepted at the IEEE ICIP Conference 2015}
\title{Efficient Image-Space Extraction and Representation of \\ 3D Surface Topography}

\name{Matthias Zeppelzauer and Markus Seidl\thanks{This work has been carried out in the project 3D-Pitoti which is funded from the European Community's 7th Framework Programme (FP7/2007-2013) under grant agreement no 600545; 2013-2016. \newline \newline Copyright 2015 IEEE. Published in the IEEE 2015 International Conference on Image Processing (ICIP 2015), scheduled for September 27-30, 2014, in Québec City, Canada. Personal use of this material is permitted. However, permission to reprint/republish this material for advertising or promotional purposes or for creating new collective works for resale or redistribution to servers or lists, or to reuse any copyrighted component of this work in other works, must be obtained from the IEEE: \newline
Manager, Copyrights and Permissions \newline IEEE Service Center, 445 Hoes Lane, P.O. Box 1331, Piscataway, NJ 08855-1331, USA\newline
Phone: +1 908-562-3966  }}
\address{Media Computing Group, Institute of Creative Media Technologies, \\ St. Poelten University of Applied Sciences, \\ Matthias-Corvinus Strasse 15, 3100 St. Poelten, Austra 
	\\  \{m.zeppelzauer\textpipe markus.seidl\}@fhstp.ac.at
	}

\begin{document}

\maketitle

\begin{abstract}
Surface topography refers to the geometric micro-structure of a surface and defines its tactile characteristics (typically in the sub-millimeter range). High-resolution 3D scanning techniques developed recently enable the 3D reconstruction of surfaces including their surface topography. In this paper, we present an efficient image-space technique for the extraction of surface topography from high-resolution 3D reconstructions. Additionally, we filter noise and enhance topographic attributes to obtain an improved representation for subsequent topography classification. Comprehensive experiments show that the our representation captures well topographic attributes and significantly improves classification performance compared to alternative 2D and 3D representations.
\end{abstract}

\section{Introduction}
\label{sec:intro}

Methods for sparse and dense 3D scene reconstruction have progressed strongly due to the availability of inexpensive, off-the-shelf hardware (e.g. Microsoft Kinect) and robust reconstruction algorithms (e.g. structure from motion techniques, SfM)~\cite{crandall2011,wu2013}. The result of these techniques is a heavily increased amount of available 3D data. Novel techniques enable dense 3D reconstructions at sub-millimeter resolution which accurately capture a surface's topography~\cite{wohlfeil2013}. This opens up new opportunities for search and retrieval in 3D scenes, such as the recognition of objects by their surface properties as well as the distinction of different types of materials for improved scene understanding. The tremendous amount of data in such high-resolution reconstructions, however, requires efficient processing methods.

The appearance of a 3D surface can be seen as the composition of the \emph{visual appearance} (visual texture, image texture) and the \emph{tactile appearance} (surface texture)~\cite{tuceryan1998, barcelo2012}. Visual appearance refers to color variations, brightness, and reflectivity. 
Surface texture refers to the geometric micro-structure of a surface in terms of roughness, waviness, and lay~\cite{blunt2003,barcelo2012}. It is defined as the repetitive random deviation from the ``ideal" surface. This deviation forms the three dimensional \emph{topography} of a surface~\cite{ansiSurfaceTexture1996}. Compared to the visual appearance of a surface, its topography is invariant to lighting and viewing conditions and thus a robust basis for its description. 

Related research has mainly focused on the description and classification of surfaces in terms of their visually apparent texture, fostered by the broad availability of image texture datasets~\cite{varma2005statistical,zhang2007,vedaldi2014texture}. Texture classification is strongly related to the description of surface topography, it is, however, highly sensitive to different lighting conditions. Thus, a major goal in texture classification is the illumination-invariant modeling of texture from one or more provided input images. In the 3D domain the situation is fundamentally different. Illumination becomes negligible, as the entire surface geometry is known. The major challenge becomes the \emph{extraction} of the surface micro-structure from 3D data and the robust \emph{description} of its attributes. 

For the representation of surface topography, descriptors are required that capture the local geometry around a given point. A large number of local 3D descriptors has been developed for this purpose, such as 3D Shape Context (3DSC)~\cite{frome_recognizing_2004} and Point Feature Histograms (PFH) \cite{rusu_fast_2009}. Their major limitation is, however, the high computational cost of their dense extraction from an entire point cloud which is necessary to capture the topography continuously across a surface. An alternative approach to represent surface topography is to extract high-frequency information from the point cloud in 3D and to map this information to 2D by non-linear dimensionality reduction \cite{othmani2013single}. The non-linear mapping is, however, a global operation on the entire point cloud which is infeasible for point clouds with several million of points. To enable the efficient extraction of topography from high-resolution reconstructions, we propose an alternative approach: Instead of extracting local 3D descriptors densely over an entire point cloud or applying global operations on the point cloud, we propose an image-space method that first maps the surface to 2D and then reconstructs and enhances the topography efficiently in the image domain. The result is a 2D topography map which is well-suited for further processing. The approach is computationally efficient and thus applicable to large-scale data.

In Section~\ref{sec:approach} we describe our approach for topography extraction. Section~\ref{sec:results} presents the experimental setup and results on a set of high-resolution surfaces. We draw conclusions and point out future work in Section~\ref{sec:conclusion}.

\section{Proposed approach}
\label{sec:approach}

We first present topography extraction from 3D data and then enhance topographic attributes for topography classification.

\subsection{Image-space topography extraction}
\label{subsec:topographyExtraction}

\begin{figure}%
	  \subfigure[original point cloud]{\label{sfig:origPC}
				\includegraphics[width=0.48\linewidth]{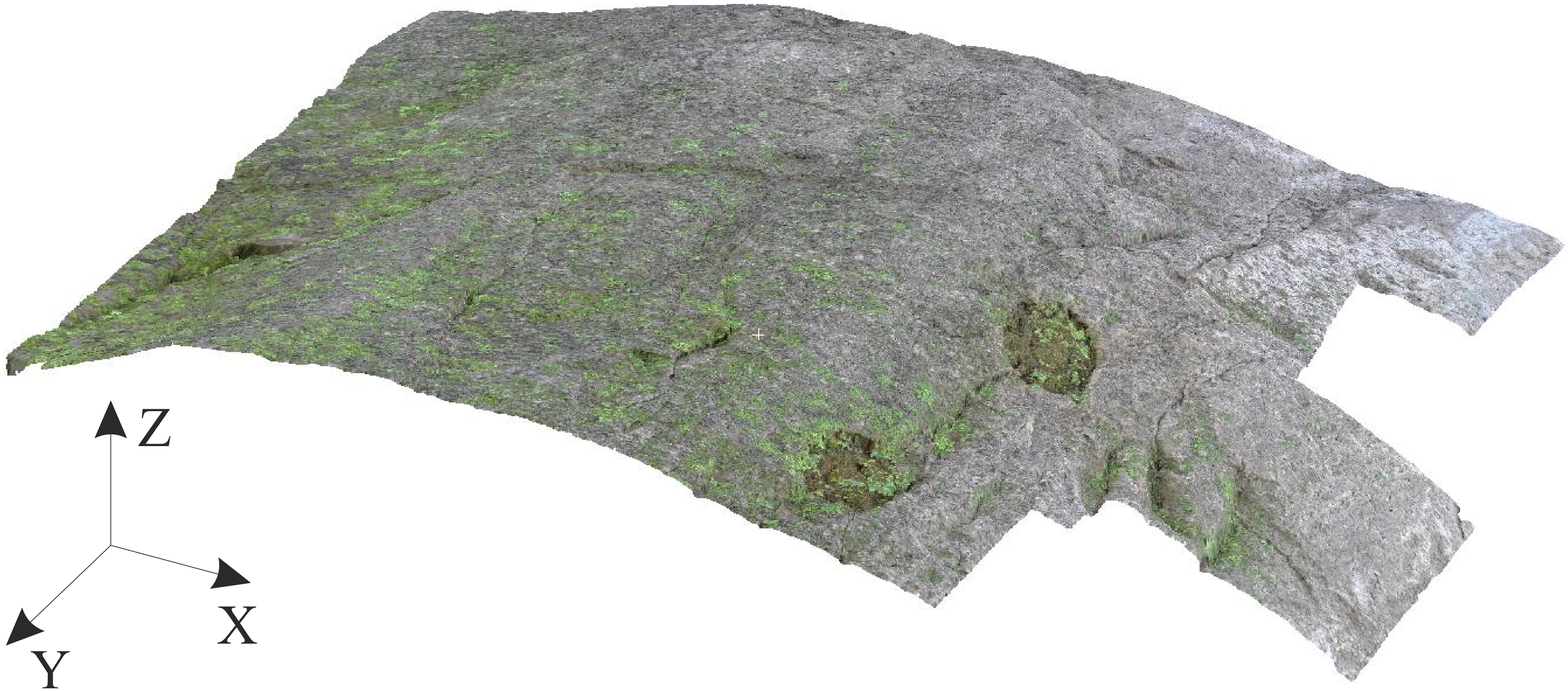}}
		\subfigure[point cloud viewed from projection direction]{\label{sfig:orthoPic}
				\includegraphics[width=0.48\linewidth]{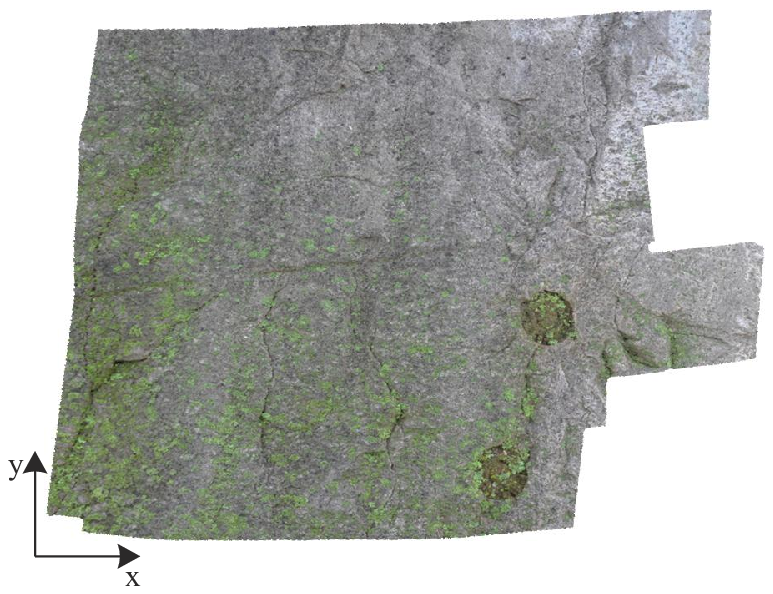}}
		\subfigure[projected depth map]{\label{sfig:depthMap}
				\includegraphics[width=0.48\linewidth]{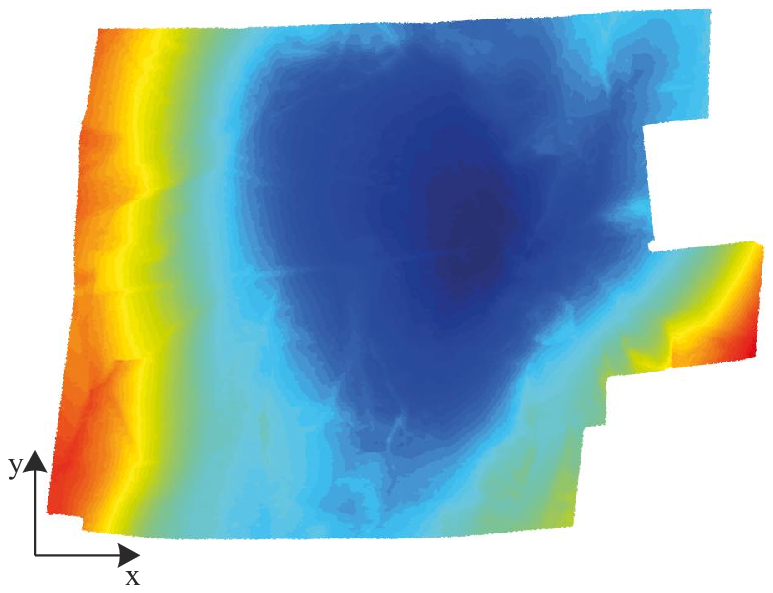}}
		\subfigure[topography map $\bar{T}$]{\label{sfig:flatDM}
				\includegraphics[width=0.48\linewidth]{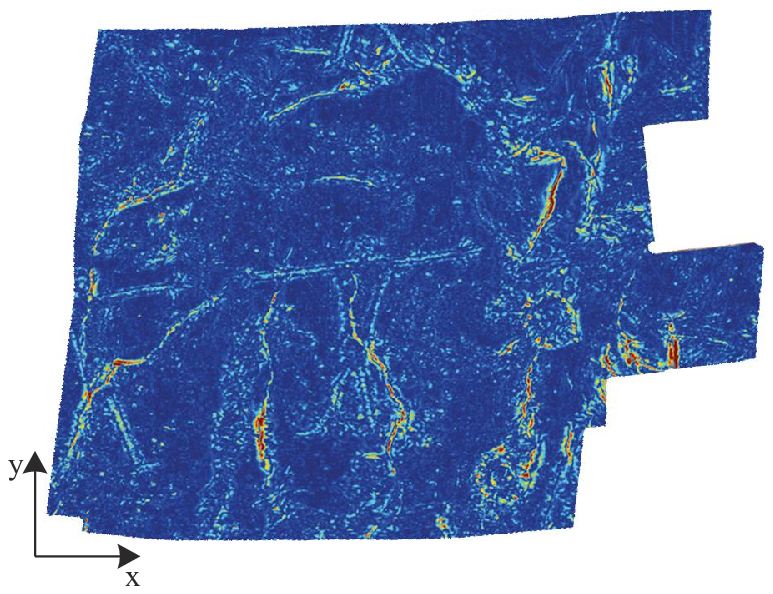}}
			\caption{Extraction of surface topography: (a) the 3D point cloud; (b) the point cloud viewed from projection direction; (c) the depth projection of the cloud; (d) the extracted topography map.}
		\label{fig:depthMap}
\end{figure}

The input to our approach is a high-resolution point cloud with $P$ points. In a first step, we estimate a support plane for the cloud that minimizes the least squares distances to the plane. Next, we estimate the location of each 3D point on the support plane by an orthographic projection with projection direction set to the normal direction of the support plane. We map the signed distances between the 3D points and the support plane to the respective locations. The result is a 2D \emph{depth map} of the 3D surface. 
Figure ~\ref{fig:depthMap} illustrates the process. Figures \ref{sfig:origPC} and \ref{sfig:orthoPic} show the original point cloud in 3D and from projection direction. The depth map is shown in Figure \ref{sfig:depthMap}. It most prominently shows the global depth variations related to the curvature of the surface. The local micro-structure is hardly visible.

The extraction of the surface topography from the depth map requires the compensation of the global curvature. We first smooth the depth map to obtain the global depth distribution of the underlying 3D surface by convolution with a two-dimensional Gaussian-shaped filter $G(x,y)$ with standard deviation $\sigma$ and a support of $W^2$ pixels. The filter is applied to all possible locations $(x,y)$ of the original depth map $D$ by convolution.
The result $\hat{D}(x,y)$ is an estimate of the local average at all surface locations. We subtract the local average from the depth map at every location by $\bar{T}(x,y) = D(x,y)-\hat{D}(x,y)$. The result is a compensated depth map $\bar{T}$, the \emph{topography map}, which is free of global geometric variations. The topography map captures well the local micro-structures of the surface (e.g. crannies and fine surface irregularities) that are not recognizable from the depth map, see Figure \ref{sfig:flatDM}. 

As most surfaces are in general not developable (non-zero Gaussian curvature) they cannot be flattened onto a plane without a certain amount of distortions. Depending on the precision required by the given application this may set a limit to the amount of global curvature contained in the surface. Since topographic attributes have local spatial support, a straight-forward approach to limit distortions is to split the surface into quasi-linear patches so that the overall curvature per patch remains limited.

The computation of the topography map is computationally efficient. The orthogonal projection can be performed in linear time $O(P)$ with respect to the number of points $P$ in the input point cloud. The convolution takes $O(Nk)$, where $N$ is the number of pixels in the projected image and $k$ the number of pixels of the kernel. The remaining operations are in $O(N)$. Thus, the method is linear in $P$ and $N$.

\subsection{Enhancement of topographic attributes}
\label{subsec:topographyEnhancement}

The topography map $\bar{T}$ is signed. The sign encodes whether a projected point is below or above the smoothed average surface and thus enables the distinction between peaks and valleys in the surface 

Peaks and valleys are important building blocks of the surface topography and are strongly related to the topographic attributes \emph{roughness} and \emph{lay}~\cite{gadelmawla2002,barcelo2012}. To obtain expressive representations for subsequent topography analysis, we extract two maps that capture the spatial distributions of peaks and valleys, respectively. Additionally, we filter noise present in the map that originates from data capturing and 3D reconstruction (sensor noise, outlier points, and noise at depth discontinuities). 
 
Positive values in the topography map indicate surface points below the smoothed surface (valleys) while negative values represent local peaks. 
We first split the topography map $\bar{T}$ into a positive and a negative part, $\bar{T}^v$ and $\bar{T}^p$ by:
\begin{equation} \label{eq:crop1}
	\bar{T}^v(x,y) = max(\bar{T}(x,y),0) 
\end{equation}
\begin{equation} \label{eq:crop2}
	\bar{T}^p(x,y) = \left|min(\bar{T}(x,y),0)\right|,
\end{equation}
where $\bar{T}^v$ contains the positive portion of $\bar{T}$ which represents the map of the local valleys and $\bar{T}^p$ captures the negative portion and provides the map of the local peaks.  
Next, we locally smooth the image with a Gaussian filter $G$ (as defined in Section~\ref{subsec:topographyExtraction}) to remove noise and discontinuities introduced by cropping in Equations \ref{eq:crop1} and \ref{eq:crop2}. As a result we obtain two filtered maps $\bar{T}^v_s = \bar{T}^v \ast G$ and $\bar{T}^p_s = \bar{T}^p \ast G$. 

In areas where the surface has strong irregularities in depth (e.g. along cracks and crannies), outliers are often obtained during 3D reconstruction. 
The outliers bias the value range of the maps and skew the distribution. To reduce their influence (and at the same time to avoid their explicit detection), we take the logarithm of both maps: $E^v = log(\bar{T}^v_s)$ and $E^p = log(\bar{T}^p_s)$. 
Logging compensates the outliers' influence and makes the value distribution approximately Gaussian. As a result the topography maps of peaks and valleys, $E^v$ and $E^p$, are improved.

\section{Experiments and results}
\label{sec:results}

We first demonstrate the effect of topography extraction and enhancement qualitatively on an example surface and then evaluate the capabilities of the proposed representations quantitatively by topography classification experiments.

\subsection{Dataset}
\label{sec:dataset}

There is currently no publicly available dataset that provides high-resolution 3D reconstructions of surfaces together with ground-truth annotations of their surface topography. A research domain that intensively analyzes surface properties is archeology. Archeologists generate high-resolution 3D reconstruction to represent findings and artifacts. For our experiments we employ a set of high-resolution 3D reconstructions of natural rock surfaces made by archeologists. The rock surfaces exhibit human-made engravings (so called rock art) which exhibit a different surface topography than the surrounding rock surface. As there additionally exists ground-truth for the engravings, the data is well-suited for topography classification and thus to evaluate the capabilities of our representations.

The employed dataset contains 4 high-resolution surface reconstructions with a total number of $12.3 \cdot 10^6$ points. The resolution is below 0.1mm in X,Y, and Z direction. For each surface a precise ground truth exists that labels the natural rock surface (class 1) and all engravings (class 2). Class 2 represents only 16.6\% of the data and is thus underrepresented.

\subsection{Qualitative evaluation}

\begin{figure}
			\subfigure[visual appearance]{\label{sfig:ortho}
				\includegraphics[width=0.45\linewidth]{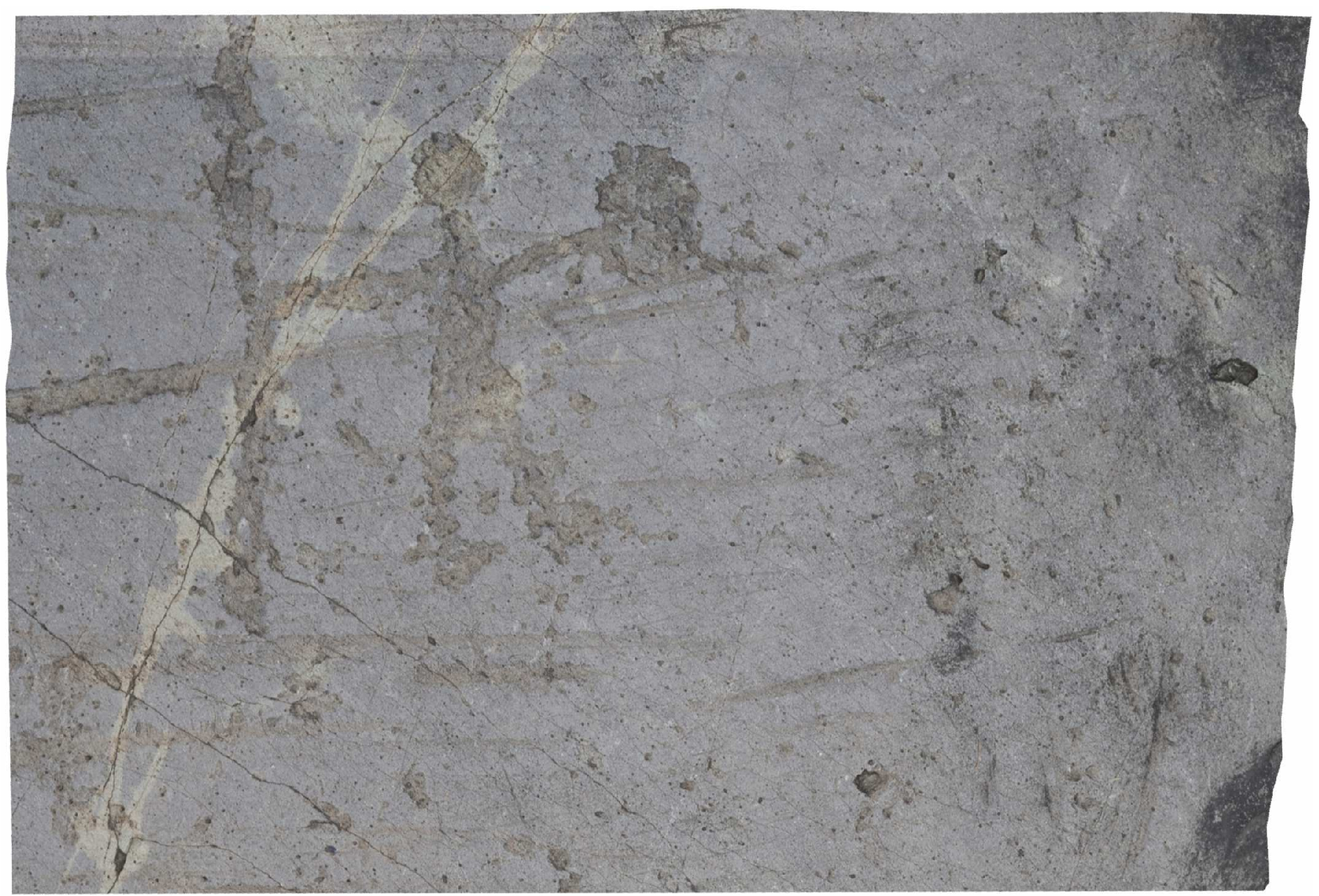}}
			\subfigure[depth map]{\label{sfig:depth_map}
				\includegraphics[width=0.45\linewidth]{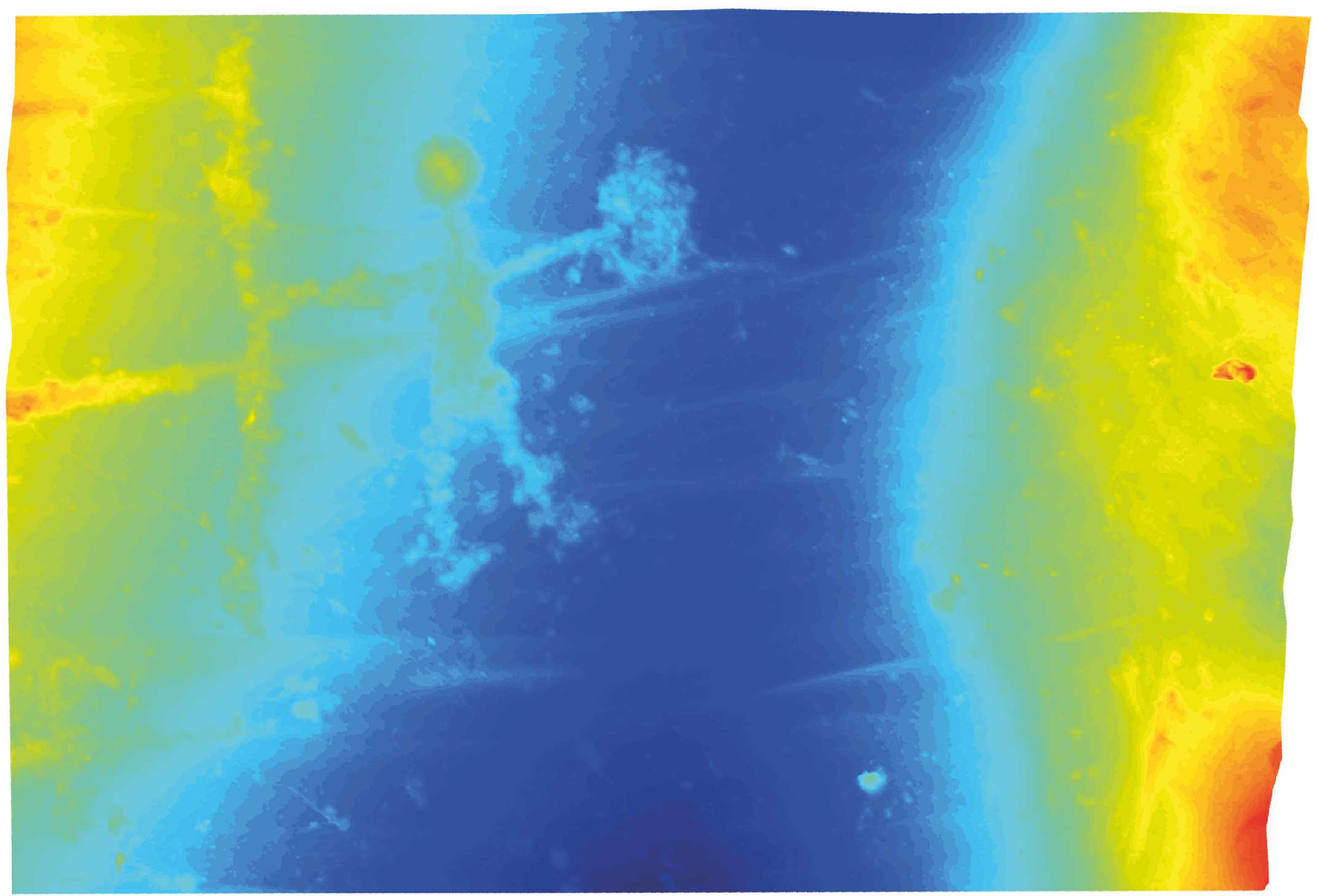}}
			\subfigure[topography map]{\label{sfig:devMap}
				\includegraphics[width=0.45\linewidth]{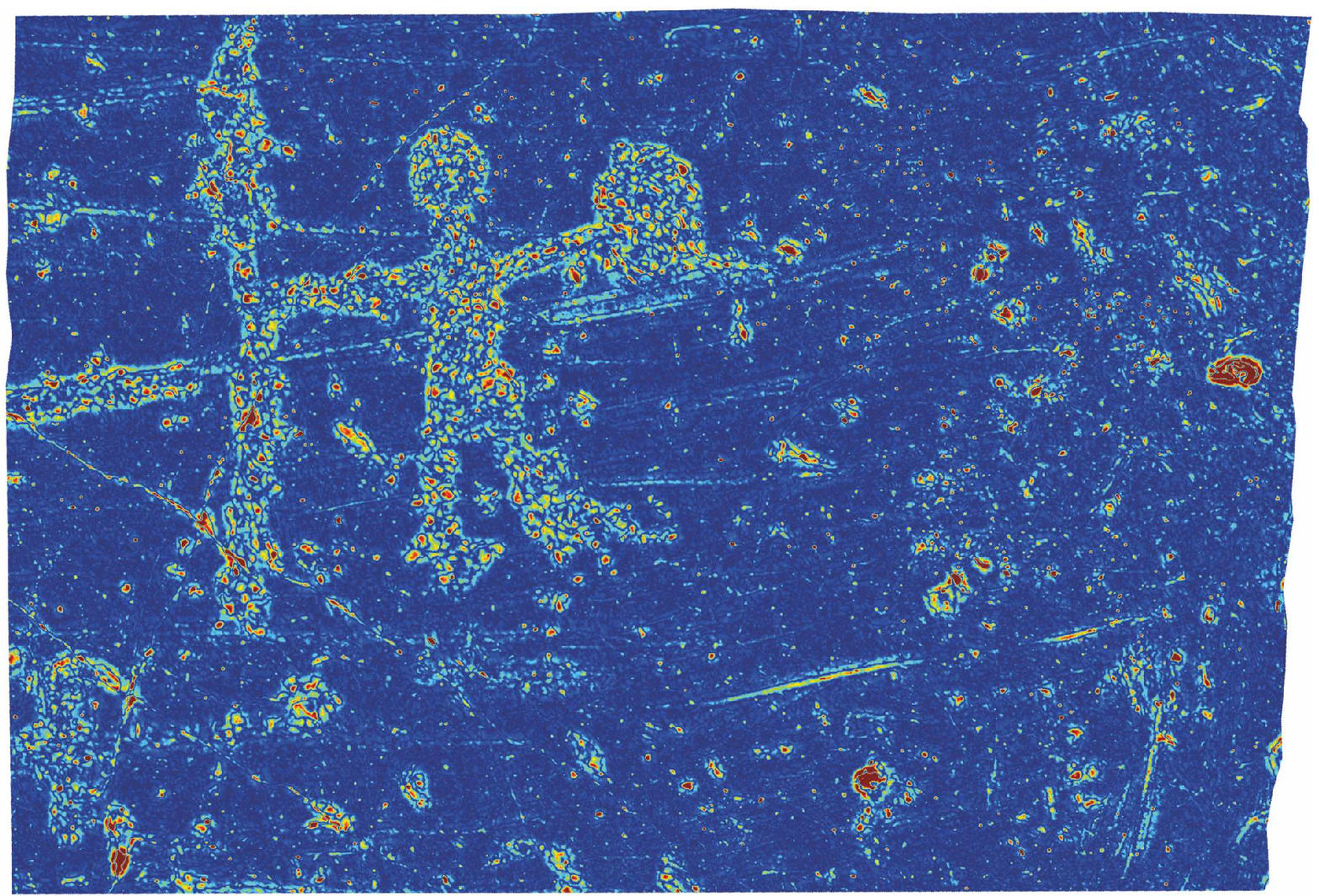}}
			\quad
			\subfigure[enhanced topography map $E^v$]{\label{sfig:final_A}
				\includegraphics[width=0.45\linewidth]{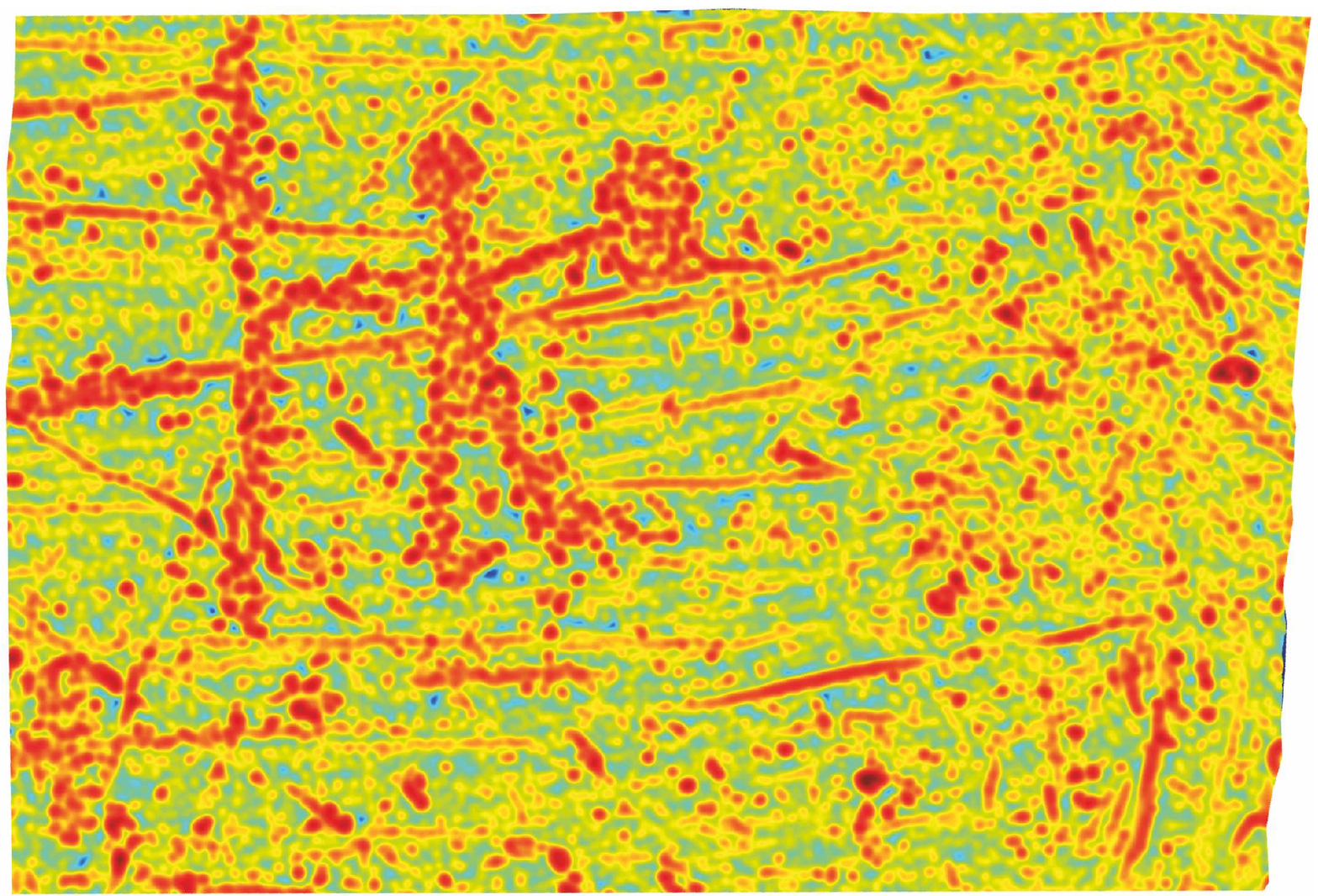}}
				
		\caption{A 3D reconstruction of a rock surface in which the shape of a human figure has been engraved. The enhanced topography map ($E^v$ in this example) accentuates well the valleys corresponding to the engravings and makes it easy to distinguish the engraving from the surrounding rock surface.} 
		\label{fig:enhancedDepthMap}
\end{figure}

We demonstrate the effect of topography extraction and enhancement on an example surface from our dataset. First, we specify the size of Gaussian filter $G$. The size of  $G$ is a design parameter that specifies at which granularity topographic structures should be enhanced. As we want to enhance topographic structures related to engravings on the rock surfaces, we set $G$ to approximately the size of an individual engraving (approx. 4mm, which refers to $W$=62 pixels). Figure \ref{fig:enhancedDepthMap} shows the result of topography extraction for a rock surface with a humanoid-shaped engraved figure. Figure \ref{sfig:ortho} shows the visual appearance (image texture) of the surface obtained with oblique shading. The depth map in Figure \ref{sfig:depth_map} clearly shows the global curvature of the surface (along the horizontal direction). At the same time, the depth map superimposes the fine surface details. The topography map in Figure \ref{sfig:devMap} compensates the global curvature of the surface and captures well the surface micro-structure. Figure \ref{sfig:final_A} shows the enhanced map $E^v$. Areas in $E^v$ with high energy (reddish color) indicate valleys in the surface. The valleys captured by $E^v$ correspond well with valleys in the surface which originate from crannies and engravings. This shows that the map models the desired topographical information. Topographic structures at finer and coarser granularity can be extracted by adapting $W$.

\subsection{Quantitative evaluation}

To evaluate the suitability of the proposed representation we perform experiments on surface topography classification using the dataset from Section \ref{sec:dataset}. As a baseline for comparison we employ (i) a full 3D appoach based on local 3D descriptors and (ii) several commonly used 2D representations as alternative to our enhanced topography maps (ETM). For the 3D approach we perform a dense analysis of the surfaces with four local 3D descriptors (PFH~\cite{rusu_persistent_2008}, FPFH~\cite{rusu_fast_2009}, SHOT~\cite{tombari_unique_2010}, and 3DSC~\cite{frome_recognizing_2004}) and apply classification directly on the 3D descriptors. 

For the 2D representations we consider the task of topography classification as a texture classification task on the respective representation. We compare the classification performance of our map with that of (i) the visual appearance of the surface (visual texture), (ii) the depth map (pure depth information), and (iii) the depth gradient map (DGM). DGM is the local gradient of the depth map. The DGM emphasizes local fine structures and neglects global depth variations and has been successfully employed in~\cite{wallenberg2011}. To capture the surface topography from the 2D representations, we extract the following baseline features (in a block-wise manner): Local Binary Patterns (LBP)~\cite{ojala2002}, SIFT descriptors~\cite{lowe2004}, Histograms of Oriented Gradients (HOG)~\cite{dalal2005}, and Gray-Level Co-occurrence Matrices (GLCM). 

Additionally, we propose two easy to compute features for the description of the ETM: (i) global histogram shape (GHS) which contains the 30 first low-frequency DCT coefficients of the global histogram of an image block and (ii) spatial frequencies (SF) which contains the first 8x8 low frequency 2D DCT coefficients of a given image block. 

For classification we employ Random Undersampling Boosting (RUSBoost) which is especially designed for imbalanced class distributions~\cite{seiffert2010}. We split the dataset into training and  test surfaces and perform 10-fold cross-validation on the training set. We compute recall and precision for both classes and report the f1 score of the \emph{underrepresented} class for each experiment which is most expressive to assess the overall performance. We apply Fisher's randomization test~\cite{fisher1935} and Student's paired t-test~\cite{box1978} with a significance level of 0.05 to judge performance differences.

\begin{figure}%
		\includegraphics[width=0.98\linewidth]{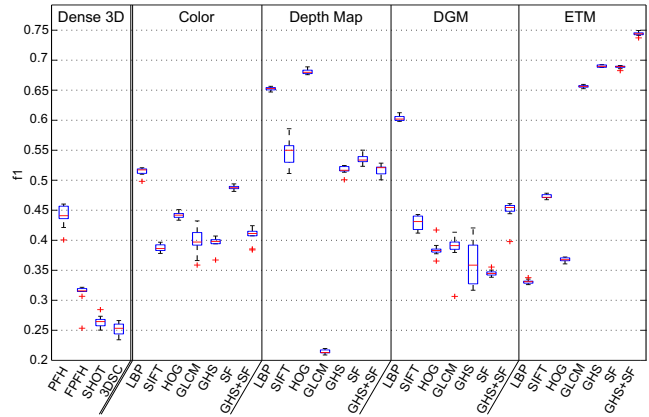}
		\caption{Results of topography classification for all 2D and 3D representations and features. Our proposed representation (ETM) significantly outperforms all other representations.}
		\label{fig:boxplots}
\end{figure}

Figure~\ref{fig:boxplots} shows boxplots of the classification results for all employed 2D and 3D representations and features. The 3D baseline obtained from dense 3D descriptors (``Dense 3D") yields weak results compared to the 2D representations (four leftmost results in Figure~\ref{fig:boxplots}) with a maximum f1 of 44.1\% for PFH. The 3D descriptors do not seem to discriminate well between the two topological classes.

Using purely visual information (image texture of the orthographic image) does also not solve the task. Results on the color image are weaker than that of the other 2D representations, with a maximum f1 score of 51.7\% for LBP. This shows that color information alone is not sufficient for the classification of surface topography. We observe that the incorporation of 3D information in terms of depth (in the depth map and in DGM) partly improves results significantly compared to color. The best results on DGM are obtained by LBP (60.2\%) and on the depth map by HOG (67.9\%), which is at the same time the best result obtained by a baseline method.

Experiments on the enhanced topography map (7 rightmost results on Figure~\ref{fig:boxplots}) show that typical texture features such as LBP and HOG cannot benefit from our representation ($f1<50\%$). GLCM, however, strongly benefits and significantly outperforms the other features evaluated so far on the ETM ($f1=65.7$). A further gain in performance is only obtained by the proposed features (GHS and SF) which both yield an f1 of 69.0\%. GHS as well as SF on ETM significantly outperform all other features (p-value$<$0.01). Peak performance in our experiments is obtained for ETM with the combination of GHS and SF ($f1=74.4\%$, p-value $<$0.01). 

We assume that the strongly varying performance of the different features on ETM has several reasons. The ETM is a comparably smooth representation (due to Gaussian filtering). Thus, features that rely on gradients, such as HOG can hardly benefit from it. Similarly LBP lacks expressiveness. GLCM benefits most, since it takes absolute values as input and does not rely on relative differences like HOG and LBP. We observe the same behavior for GHS and SF which rely on absolute values, as well. The high performance of the combined features (GHS+SF) has two reasons: firstly, both features perform well when applied separately; secondly, both features represent complementary information: GHS represents the value distribution globally while SF captures the spatial value distribution.

\section{Conclusion}
\label{sec:conclusion}

We have presented an efficient image-space method for the extraction of surface topography from high-resolution 3D reconstructions. Additionally, we have proposed enhanced topography maps as a novel representation for improved topography classification. Our evaluation demonstrates that the proposed maps are well-suited for topography representation and significantly outperform alternative representations in classification experiments, such as local 3D descriptors, depth maps, and depth gradient maps. 
Furthermore, experiments show that pure 2D information (image texture) is not able to capture the full surface topography. In future, we will extend topography maps to multiple scales, to obtain a representation that captures topographic structures at different granularities.

\bibliographystyle{IEEEbib}
\bibliography{refs}

\end{document}